\documentclass{article}

\usepackage{arxiv}
\usepackage[round]{natbib}

\usepackage[utf8]{inputenc}
\usepackage{graphicx,times}
\usepackage{amsmath,amsthm,amssymb,amsfonts,latexsym}

\usepackage{algorithm}%
\usepackage{algpseudocode}%

\usepackage{bbm}
\usepackage{bm}
\usepackage{paralist}
\usepackage{qtree}

\usepackage{multirow}
\usepackage[normalem]{ulem}
\usepackage{makecell}

\newcommand\comment[1]{}

\def\l[{\left[}
\def\r]{\right]}

\def\Xs{\mathcal{X}}
\def\Ys{\mathcal{Y}}

\def\ie{\emph{i.e., }}
\def\eg{\emph{e.g., }}

\title{Improving Classifier Confidence using Lossy Label-Invariant Transformations}

\author{ Sooyong Jang\\PRECISE Center\\University of Pennsylvania
\And Insup Lee\\PRECISE Center\\University of Pennsylvania
\And James Weimer\\PRECISE Center\\University of Pennsylvania }
\date {}
\begin{document}
\maketitle

\begin{abstract}

Providing reliable model uncertainty estimates is imperative to enabling robust decision making by autonomous agents and humans alike.  While recently there have been significant advances in confidence calibration for trained models, examples with poor calibration persist in most calibrated models.  Consequently, multiple techniques have been proposed that leverage label-invariant transformations of the input (\ie an input manifold) to improve worst-case confidence calibration.  However, manifold-based confidence calibration techniques generally do not scale and/or require expensive retraining when applied to models with large input spaces (\eg ImageNet).  In this paper, we present the recursive lossy label-invariant calibration (ReCal) technique that leverages label-invariant transformations of the input that induce a loss of discriminatory information to recursively group (and calibrate) inputs -- without requiring model retraining.  We show that ReCal outperforms other calibration methods on multiple datasets, especially, on large-scale datasets such as ImageNet.

\end{abstract}

\section{Introduction}
\label{sec:introduction}
\vspace{-1em}

Despite the success of machine learning predictions in various applications
including image classifications \citep{he2016deep, zagoruyko2016wide, xie2017aggregated}, speech recognition \citep{graves2013speech, wang2019espresso}, games \citep{mnih2013playing, silver2017mastering}, and medical research \citep{rajpurkar2018deep},
estimating prediction confidence has a different story.
As observed in \cite{guo2017calibration}, many modern neural networks are miscalibrated, \ie 
they are over-confident in their predictions.
As machine learning expands to safety-critical applications such as self-driving cars, autonomous pilots, and autonomous medical systems, accurately estimating confidence becomes imperative for robust decision making.

Consequently, various approaches have been introduced to address the problem of estimating confidence.  Bayesian techniques \citep{gal2016dropout, zhang2017noisy, khan2018fast, chang2019ensemble} provide a means of computing the posterior distribution of models for estimating confidence, but suffer from computational limitations.  
Also proposed are techniques that change the original model estimates~\citep{tran2019calibrating,kumar2018trainable, pereyra2017regularizing, seo2019learning}, but these techniques have the disadvantage that they require re-training the model and do not guarantee the accuracy of the original model. 
Lastly, there have been many post-hoc approaches proposed that learn a model mapping uncalibrated confidence to calibrated confidence on a comparatively small validation set
\eg temperature scaling, vector scaling\citep{guo2017calibration, rahimi2020intra}, using spline \citep{gupta2020calibration}, MS-ODIR and Dir-ODIR \citep{kull2019beyond}, mix-n-match \citep{zhang2020mix},  and GPcalib \citep{wenger2020non}. While these techniques provide improved average confidence calibration, poorly calibrated examples remain.

To address this issue, techniques that utilize redundancy in the example space have been proposed \citep{bahat2020classification, thulasidasan2019mixup, patel2019manifold}.
The premise behind these techniques is that utilizing additional information on the current sample, its confidence calibration can be improved.  Most of these techniques augment the training dataset with examples on the same manifold and re-train a model \citep{thulasidasan2019mixup, patel2019manifold} on the augmented dataset.  While other techniques, such as \citep{bahat2020classification}, avoid retraining by assuming there exist well calibrated examples within the manifold and perform filtering of a sampling of the manifold confidences.
While evaluating the confidence of an individual sample 
remains a challenge -- since most benchmark datasets do not contain confidence labels (only class labels) -- these techniques do generally show marked improvement in expected confidence calibration consistent with a reduced number of poorly calibrated samples.  However, manifold-based techniques have severe shortcomings when applied to datasets with large input spaces.  Augmenting the training dataset and retraining a model scales poorly as the input space (and manifold dimensionality) increases.   Similarly, filtering sampled confidences assumes that the original classifier is calibrated for a majority of the manifold, which becomes less likely as manifold dimensionality increases.  Consequently, manifold-based calibration of models for large input spaces remains a challenge.

\begin{figure}[!bt]
     \centerline{
    \includegraphics[width=0.35\textwidth]{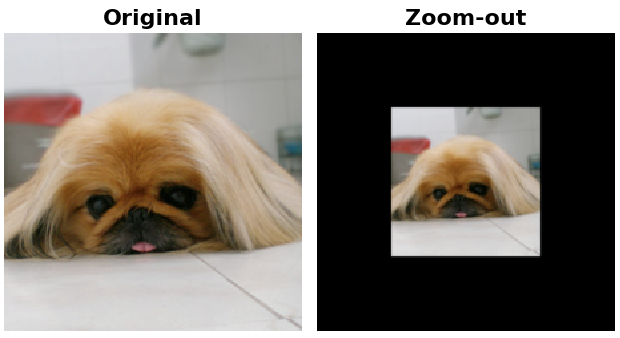}
    }
    \caption{Example of Original and Transformed Image} 
    \label{fig:sample_img}
\end{figure}

In this work, we present \emph{Recursive Lossy Label-Invariant Calibration} (or ReCal) as a scalable manifold-based 
post-hoc confidence calibration algorithm that maintains the accuracy of the original classifier and scales to large datasets (\emph{e.g.} ImageNet).  
To overcome the scalability issues 
of other manifold-based techniques, we only consider label-invariant transformations that are expected to result in a decreased confidence due to discriminatory information loss -- \ie lossy label-invariant transformations.
For example, consider zooming out an image of a dog with the scale factor of 0.5x as shown in Figure \ref{fig:sample_img}.
After the transformation, the image still contains the dog, but the dog becomes smaller and harder to recognize.
Therefore, we should be able to identify the dog but with less confidence. 
Considering this intuition in the context of estimating confidence, a (well-calibrated) classifier should return the same prediction with smaller confidence after applying a lossy label-invariant transformation.
Likewise, if we group examples based on the prediction and confidence change after such transformations, we expect that the examples in the same group will have similar properties respect to the classifier and confidence estimation.
In other words, examples in each group require a similar amount of calibration, which may be different than the calibration needed for examples in other groups. 
This intuition -- lossy label-invariant grouping -- forms the premise of ReCal, and is discussed in detail in Section \ref{sec:group_input}.

Leveraging group-wise calibration, we propose ReCal as a scalable post-hoc calibration algorithm in Section \ref{sec:algorithm}. Specifically, the proposed algorithm recursively leverages lossy label-invariant transformations to re-group images and perform group-wise calibration. Different from other approaches that aim to retrain a model on the augmented training set ~\citep{patel2019manifold, thulasidasan2019mixup}, our proposed algorithm does not change the predictions and thus retains the original prediction accuracy while adjusting the confidence of the predictions.

We demonstrate the scalability and performance of the proposed algorithm by applying it to ImageNet, and also compare ReCal with other calibration algorithms on CIFAR10/100, ImageNet in Section \ref{sec:experiments}.
On multiple models \eg LeNet5, DenseNet, ReseNet, ResNet SD, and Wide ResNet,  on the datasets, we compare Expected Calibration Error (ECE) \citep{naeini2015obtaining}, Brier score \citep{brier1950verification} and time for learning a calibration map. 
On the large scale image dataset, ImageNet, ReCal can be applied to the dataset in terms of time, and it outperforms other calibration algorithms such as temperature scaling, vector scaling \citep{guo2017calibration}, MS-ODIR, Dir-ODIR\citep{kull2019beyond}
on DenseNet161 and ResNet152 models. %
Besides ImageNet, ReCal shows the best performance or the second-best performance for seven of ten models on CIFAR10/100. 

The contributions of this paper are summarized as follows:
\begin{itemize}
\vspace{-1em}
\item introducing lossy label-invariant grouping and empirically demonstrating that each group needs different calibration; \vspace{-.5em}
\item presenting ReCal, a scalable post-hoc calibration algorithm based on lossy label-invariant transformation, which can be applied to a large-scale datasets; \vspace{-.5em}
\item evaluating ReCal in comparison to other publicly released post-hoc calibration algorithms using multiple datasets and models. \vspace{-.5em}
\end{itemize}
\vspace{-.5em}
The remainder of this paper is structured as follows.  In the next section, we present the related work on confidence calibration.  In Section \ref{sec:problem_statement}, we present the problem statement considered herein. Section \ref{sec:group_input} describes lossy label-invariant grouping and its effectiveness. We then propose ReCal in Section \ref{sec:algorithm}, present the experimental results in Section \ref{sec:experiments}, and conclusions in Section \ref{sec:conclusion}.

\section{Related Work}
\label{sec:related_work}
\vspace{-1em}

While a complete review of all confidence calibration techniques is beyond the scope of this work, in this section we selectively review those techniques most related to the proposed approach.  
In the following, we consider confidence calibration techniques leveraging Bayesian uncertainty estimation, calibration via re-training, post-hoc calibration maps, and manifold-based calibration.

\textbf{Bayesian uncertainty estimation.}
One approach to confidence calibration is to provide uncertainty estimation with Bayesian framework \citep{gal2016dropout, zhang2017noisy, khan2018fast, chang2019ensemble}. While Baysian techniques can provide very accurate calibration, they suffer from computational limitations associated with estimating the posterior distribution used for uncertainty estimation.

\textbf{Calibration via re-training}
Another type of approach targets training a well-calibrated classifier \citep{kumar2018trainable, lakshminarayanan2017simple, pereyra2017regularizing, seo2019learning, tran2019calibrating, muller2019does}.  A potential pitfall of calibration via re-training is that the accuracy of the prediction can change.  Moreover, many of these approaches require training sophisticated networks on large training datasets, which may consume significant time and computational resources. 

\textbf{Post-hoc calibration maps.}
Post-hoc methods address the calibration problem without requiring model retraining.  These approaches employ binning methods such as Histogram Binning \citep{guo2017calibration}, Bayesian Binning into Quantiles \citep{naeini2015obtaining}, Mutual Information Maximization-based Binning \citep{patel2020multi} or train a function mapping from original confidence to calibrated one on validation data which is smaller compared to the training data.
For training a mapping function, several techniques have been proposed \citep{platt1999probabilistic, zadrozny2002transforming, guo2017calibration, rahimi2020intra, gupta2020calibration, kull2019beyond, zhang2020mix, wenger2020non}. Most notably,  \cite{guo2017calibration}, introduces temperature scaling which transforms original logits to calibrated logits with a single parameter. 
Besides temperature scaling, intra-order preserving function \citep{rahimi2020intra}, Dirichlet calibration with ODIR regularization \citep{kull2019beyond}, splines \citep{gupta2020calibration}, latent Gaussian function (GPcalib) \citep{wenger2020non}, and ETS, IRM, IROvA-TS \citep{zhang2020mix} have been proposed.
Depending on the mapping function, some of the approaches such as temperature scaling, intra-order preserving function, splines, ETS and IRM preserve the accuracy, while the others like matrix scaling, vector scaling, IROvA-TS, GPcalib and Dirichlet calibration do not preserve the original model accuracy.

\textbf{Manifold-based calibration.}
Several manifold-based confidence calibration have been proposed \citep{bahat2020classification, thulasidasan2019mixup, patel2019manifold,lee2017training,verma2019manifold}. 
\cite{bahat2020classification} augments test data using transformations to calibrate confidence, while 
\cite{thulasidasan2019mixup} and \cite{patel2019manifold} augment data by interpolating existing data and using an auto-encoder based model, respectively.  Other techniques augment the training data with samples from the manifold and retrain the model~\citep{lee2017training,verma2019manifold}.
Manifold-based algorithms can improve worst-case calibration errors as shown by their ability to address over-confident prediction on out-of-distribution samples.  However, they generally suffer from scalability issues as discussed in Section \ref{sec:introduction}.

\section {Problem Statement}
\label{sec:problem_statement}
\vspace{-1em}

In this paper, we aim to develop a post-hoc calibration algorithm which addresses the worst-case confidence error that does not change accuracy on a multi-class classification task.
Consider a multi-class classification task on data, $D = \{(x_n, y_n)\}^N \sim \Xs \times \Ys$, where $\Xs$ is an input space and $\Ys$ is a label set, $\{1, 2, \dots, K\}$.
Let $f: \Xs \to \mathbb{R}^K$ denote a multi-class classifier.
A neural network classifier typically has a softmax output layer as a final layer, which returns a vector $z$ for the given input $x$.
Here, $z = f(x) =\{z_1, z_2, \dots, z_k\}$.
Each $z_i$ is the estimated probability of a label $i$, and the classifier chooses the label whose probability is the maximum.
Consequently, the prediction $\hat{y}=\text{argmax}_i\{z\}$ has confidence $p = \text{max}_i\{z\}$.

A classifier $f$ is \textit{calibrated} if confidence is equal to accuracy given the confidence.
More formally,
\begin{equation}
    \mathbf{P}[y = k | z_k = p] = p
\end{equation}
where, $z = f(x), k = \text{argmax}_i\{z\}$ for all $(x, y) \in D$ and for all $p \in [0, 1]$.
Here, the difference between the both sides is defined as \textit{calibration error}.
This error is estimated by \textit{Expected Calibration Error (ECE)} \citep{naeini2015obtaining} which is the weighted average of the differences over bins.
ECE is computed by first splitting the confidence range with equal size bins, and calculating the difference between average confidence and average accuracy of each bins, and finally, computing an average of those differences weighted by the number of samples in each bins.
More formally,
\begin{equation}
\label{eq:ece_def}
    \text{ECE} = \sum_{i=1}^{N}\frac{|B_i|}{|D|}\left(|accuracy(B_i) - confidence(B_i)|\right)
\end{equation}
where, $N$ is the number of bins, $B_1, \dots, B_N$ are bins which equally divides the interval [0, 1], and $accuracy(B_i)$ is the average accuracy of examples in bin $B_i$, and $confidence(B_i)$ is the average confidence of examples in bin $B_i$.

Ideally, we would like to minimize the worst-case confidence error, however, this is impossible to quantify with the current datasets that lack ground truth calibration values for the labels.
As a surrogate, and consistent with other works in the literature, we rather aim to minimize ECE. Therefore, we would like to learn a calibration map which transforms the original confidence (or logits) to calibrated one which minimize ECE without affecting original accuracy.

\section{Lossy Label-Invariant Grouping}
\label{sec:group_input}
\vspace{-1em}

\begin{figure}[!t]
    \centerline{
    \includegraphics[width=.4\textwidth]{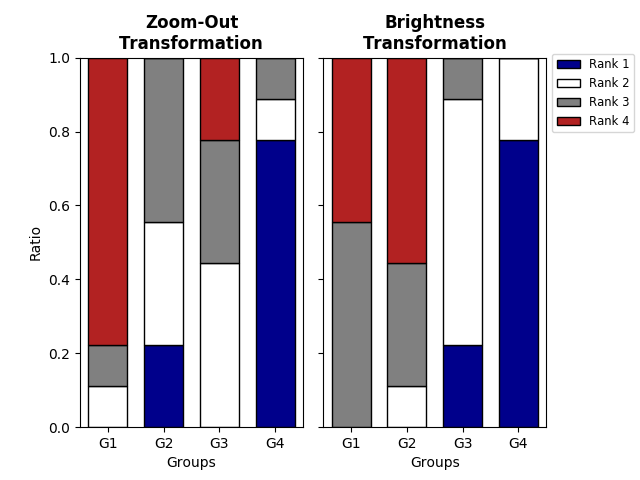}
    }
    \caption{Rank Distribution of Each Group with Two Transformations. Each Bar Represents Rank Distribution of Each Group over Different Transformation Parameters. }
    \label{fig:onetime}
\end{figure}

It is reasonable to assume that different examples may need different level of calibration, \ie some examples require more calibration than others.
Consequently, we would like to group inputs based on some measure of calibration needed so that we can apply different level of calibration to each group.
In other words, we would like to apply more calibration when predictions are very mis-calibrated, and calibrate less when predictions near calibration.

We utilize a subset of transformations that do not change the label called \emph{label-invariant transformations}.
Specifically, we choose label-invariant transformations which induce a loss in discriminatory information -- \ie lossy label-invariant transformations.
As an example, consider an image classification task.  The zoom-out transformation and brightness transformation are the examples of lossy label-invariant transformations. 
These two transformation do not change label, but reduce discriminatory information by making an image smaller or darker.
Therefore, after such transformations, a (well-calibrated) classifier should not change its prediction but should become less confident on its prediction.

Our approach, as described in Algorithm \ref{alg:group_input}, begins by applying a lossy label-invariant transformation to the inputs, and group based on the observed prediction and confidence changes after the transformation.
There can be two possible outcomes to each observation, \ie prediction change vs not change, and confidence increase vs. not increase, and in total there can be four possible combinations as shown in Table \ref{tab:input_group}. 
We perform \emph{lossy label-invariant grouping} by comparing the prediction and confidence of the transformed input with the original input, and group based on the comparison result.  
More formally, group number $k$ for an input is
\begin{align}
    k = 2  \times \mathbbm{1}_{(\hat{y} = \hat{y}_t)} +  \mathbbm{1}_{(\hat{p} \ge \hat{p}_t)} + 1
\end{align}

where, $\hat{y}, \hat{p}$ are the prediction and confidence for the original input, $\hat{y}_t, \hat{p}_t$ are the prediction and confidence for the transformed input, and $\mathbbm{1}_{(\cdot)}$ is 1 if $(\cdot)$ is true, and 0, otherwise.

\begin{algorithm}[!t]
\caption{Lossy Label-Invariant Grouping}\label{alg:group_input}
\begin{algorithmic}[1]
\Procedure{grp\_input}{$z, z_t$}
    \State $\textbf{Input:}$ $z:$ $(N_{v} \times C)$ Original inputs logits; $z_t:$ $(N_{v} \times C)$ Transformed inputs logits

    \State $\hat{y} \gets$ argmax($z$, axis=1)
    \State $\hat{y}_t \gets$ argmax($z_t$, axis=1)
    \State $\hat{p} \gets z^{\hat{y}}$
    \State $\hat{p}_t \gets z_t^{\hat{y}}$
    \State $g\_1\_idx = (\hat{y} \neq \hat{y}_t ) \wedge (\hat{p}_t > \hat{p})$
    \State $g\_2\_idx = (\hat{y} \neq \hat{y}_t ) \wedge (\hat{p}_t \le \hat{p})$
    \State $g\_3\_idx = (\hat{y} = \hat{y}_t ) \wedge (\hat{p}_t > \hat{p})$
    \State $g\_4\_idx = (\hat{y} = \hat{y}_t ) \wedge (\hat{p}_t \le \hat{p})$
    
    \State $\textbf{return } g\_1\_idx, g\_2\_idx, g\_3\_idx, g\_4\_idx$ 
\EndProcedure
\end{algorithmic}
\end{algorithm}

To demonstrate the effectiveness of our Lossy Label-Invariant Grouping Algorithm we consider an image classification task using ResNet152 model on ImageNet.
We choose two image transformations, zoom-out and brightness.
These two transformations have one parameter which determines how much transformation will be applied.
A zoom-out transformation with smaller parameter value will return the smaller image and a brightness transformation with smaller parameter value will yield the darker image. 

We randomly select parameter values between 0.1 and 0.9, we observe label prediction change and confidence change for validation images, and group images into four different groups as shown in Table \ref{tab:input_group}.
For each parameter, we compute and rank the ECE values among four groups, and draw the distribution of the ranks 
for each transformation as shown in Figure \ref{fig:onetime} .
For the reference, ECE values and number of images of each group for the two transformations are displayed in the supplementary material.

\begin{table}[!t]
    \caption{Grouping Inputs Based On Prediction and Confidence Change. $\hat{y}, \hat{p}$ Are the Prediction and Confidence for Original Input, and $\hat{y}_t, \hat{p}_t$ Are the Prediction and Confidence for Transformed Input.}
    \label{tab:input_group}
    \begin{center}
    \begin{tabular}{c|c|c}
         & $\hat{p} < \hat{p}_t$ & $\hat{p} \ge \hat{p}_t$ \\
    \hline
     $\hat{y} \neq \hat{y}$ & Group 1 & Group 2 \\
     $\hat{y} = \hat{y}$  & Group 3 & Group 4\\
    \end{tabular}
    \end{center}
\end{table}

As shown in Figure \ref{fig:onetime} (Left), with zoom-out transformation, for about 80\% of parameters, Group 4 which represents that prediction does not change and confidence does not increase, has the best ECE, \ie rank one.
On the other hand, Group 1 which corresponds to the case that prediction changes and confidence increases shows the worst ECE, \ie rank four.
The similar pattern appears on the brightness transformation as shown in Figure \ref{fig:onetime} (Right).
With brightness transformation, Group 4 has the best ECE for about 80\% of parameters, and Group 1 has the worst ECE for about 50\% of parameters.
Group 2 and Group 3 show a little different pattern between two transformations.
For zoom-out transformation, Group 2 has better rank than Group 3 with about 80\% of each group is either rank 2 or 3.
On the other hand, for brightness transformation, Group 2 has worse rank than Group 3.
It is hard to decide which one requires more calibration, but in general, these two groups should be calibrated differently. 
\begin{figure*}[!t]
    \centerline{
    \includegraphics[width=1\textwidth]{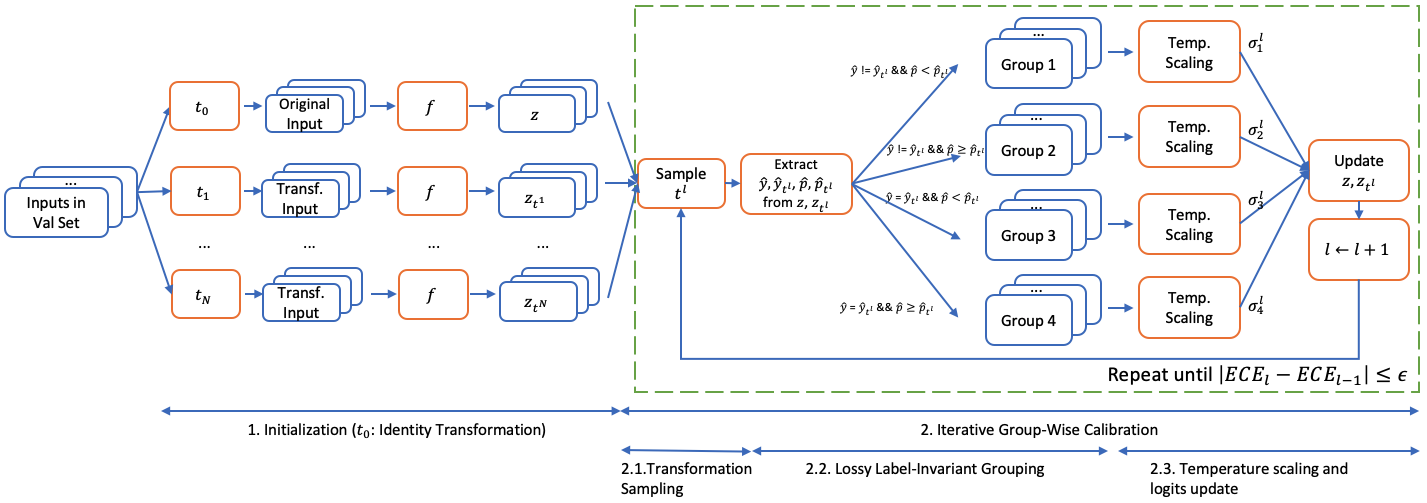}
    }
    \caption{Illustration of Recursive Lossy Label-Invaraint Calibration (ReCal)} 
    \label{fig:flow_chart_cal}
\end{figure*}

These results empirically demonstrate that lossy label-invariant grouping partitions the inputs into groups that require different amounts of calibration.
Group 4 inputs which match our intuition tend to have the best ECE, \ie requires the least calibration, while Group 1 inputs which opposite to our intuition show the worst ECE, \ie requires the most calibration. 
Furthermore, input grouping differs depending on the transformation, as shown by the input distribution over groups for different transformations in the supplementary Tables.
Consequently, in the following section, we design an algorithm that utilizes different lossy label-invariant transformations at each iteration to perturb the groupings and perform recursive calibration.

\section{Recursive Lossy Label-Invariant Calibration (ReCal)}
\label{sec:algorithm}
\vspace{-1em}

\begin{algorithm*}[!hbt]
\caption{Recursive Lossy Label-Invariant Calibration (ReCal)}\label{alg:iter_cal}
\begin{algorithmic}[1]
\Procedure{re\_cal}{$ts, X, y, N, L, \delta$}
    \State \textbf{Input:} $ts$: $(N_{allow} \times 1)$ Transformations specification; $X$: $(N_{v} \times p)$. Inputs in validation set; $y$:$(N_{v} \times 1)$. True labels; $N$:$(1 \times 1)$. Number of transformations; $L$:$(1 \times 1)$. Maximum iteration number; $\delta$:$(1 \times 1)$. Stopping iterations threshold
    
    \State $\{t_1, \dots t_N\} \gets$ Build a transformation pool based on $ts$
    \State $z \gets$ Base logits for original inputs 
    \State $z_{t_1}, \dots z_{t_N} \gets$ Base logits for transformed inputs
    
    \For {$l=1,2,\ldots, L$}
        \State $t^l \gets$ Randomly select a transformation from $\{t_1, \dots t_N\}$
        \State Group logits $z$ and $z_{t^l}$ using $grp\_img$ in Algorithm \ref{alg:group_input}
        \State Apply temperature scaling to group 1 - 4 and obtain temperature parameters, $\hat{\sigma}^l_1, \hat{\sigma}^l_2, \hat{\sigma}^l_3, \hat{\sigma}^l_4, $
        \State Compute temperature parameters $\sigma^l_1, \sigma^l_2, \sigma^l_3, \sigma^l_4$ using Equation \ref{eq:update_temp} 
        \State Calibrate logits for original inputs, $z^1, z^2, z^3, z^4$ and logits for transformed inputs, $z^1_{t^l}, z^2_{t^l}, z^3_{t^l}, z^4_{t^l}$
        \State Update logits for original inputs $z$ using $z^1, z^2, z^3, z^4$, and logits for transformed inputs $z_{t^l}$ using $z^1_{t^l}, z^2_{t^l}, z^3_{t^l}, z^4_{t^l}$
        \If {$|ECE_l - ECE_{l-1}| < \delta$}
            \State $\textbf{return } \sigma^1_1, \sigma^1_2, \sigma^1_3, \sigma^1_4, \dots, \sigma^l_1, \sigma^l_2, \sigma^l_3, \sigma^l_4, t^1, \dots, t^l, l $ 
        \EndIf
    \EndFor
\EndProcedure
\end{algorithmic}
\end{algorithm*}

As illustrated in Figure \ref{fig:flow_chart_cal} and Algorithm \ref{alg:iter_cal}, ReCal consists of 3 steps: initialization, iterative group-wise calibration, final calibration.
In the following we describe each of these steps in detail.  We conclude this section by presenting a runtime implementation of ReCal and a discussion of its limitations.

\subsection{Initialization}
\vspace{-.5em}
To initialize ReCal, a transformation pool is prepared by sampling $N$ transformations $\{t_1, \dots, t_N\}$ from possible transformations.
After a transformation pool is prepared, ReCal computes base logits of the original inputs and transformed inputs.
The logits of the original inputs are obtained by feeding the original inputs to the original confidence estimator.
For the logits of the transformed inputs, the original inputs are transformed by the sampled $N$ transformations, and fed to the same original confidence estimator.

\subsection{Iterative Group-Wise Calibration}
\vspace{-.5em}
The following three steps will be repeated up to the maximum iteration, $L$, or until the stopping condition is satisfied: (i) Transformation sampling; (ii) Lossy label-invariant grouping; (iii) Temperature scaling and logits update. This algorithm is described in Line 7 - 15 of Algorithm \ref{alg:iter_cal} and each step is detailed below.

{\bf \emph{Transformation sampling}.} 
At each iteration $l$, a transformation $t^l$ is randomly sampled with replacement from a transformation pool $\{t_1, \dots, t_N\}$.

{\bf \emph{Lossy label-invariant grouping}.} 
Once a transformation $t^l$ is sampled, inputs in a validation set will be grouped using the lossy label-invariant grouping algorithm presented in Section \ref{sec:group_input}.

\begin{algorithm*}[!bt]
\caption{Runtime Confidence Calculation using ReCal}\label{alg:apply_cal}
\begin{algorithmic}[1]
\Procedure{apply\_cal}{$X,  \sigma^1_1, \sigma^1_2, \sigma^1_3, \sigma^1_4, \dots, \sigma^{L^*}_1, \sigma^{L^*}_2, \sigma^{L^*}_3, \sigma^{L^*}_4,  t^1, \dots, t^{L^*}, L^*$}
    \State \textbf{Input:} $X$: $(N_{te} \times p)$. Test set inputs; $\sigma^1_1, \sigma^1_2, \sigma^1_3, \sigma^1_4, \dots, \sigma^{L^*}_1, \sigma^{L^*}_2, \sigma^{L^*}_3, \sigma^{L^*}_4: (4L^* \times 1)$. Temperature parameters ;$t^1, \dots, t^{L^*}$: $(L^* \times 1)$. Sampled transformation for each iteration; $L^*$:$(1 \times 1)$. Iteration number
    
    \State $z \gets$ Base logits for original inputs
    \State $z_{t_1}, \dots z_{t_N} \gets$ Base logits for transformed inputs
    
    \For {$l=1,2,\ldots, l^*$}
        \State Calibrate original inputs logits, $z^1, z^2, z^3, z^4$
        \State Calibrate transformed inputs logits, $z^1_{t^l}, z^2_{t^l}, z^3_{t^l}, z^4_{t^l}$ 
        \State Update original inputs logits $z$ using $z^1, z^2, z^3, z^4$
        \State Update transformed inputs logits $z_{t^l}$ using $z^1_{t^l}, z^2_{t^l}, z^3_{t^l}, z^4_{t^l}$
    \EndFor
    
    \State \textbf{return} updated logits for original inputs
\EndProcedure
\end{algorithmic}
\end{algorithm*}
{\bf \emph{Temperature scaling and logits update}.}
For each lossy label-invariant group, temperature scaling \citep{guo2017calibration} is applied, and temperature parameters, $\hat{\sigma}^l_1, \hat{\sigma}^l_2, \hat{\sigma}^l_3, \hat{\sigma}^l_4$, are generated -- one corresponding to each group.
Each group will have different number of inputs and overfitting can occur if the number of inputs is small.  
To safeguard against overfitting, we modify the temperature parameter based on the number of inputs in the group as shown in Equation \ref{eq:update_temp}.
\begin{equation}
\label{eq:update_temp}
\sigma^l_k = \left(1-\frac{|G_k|}{|D_{val}|}\right) \times 1 + \frac{|G_k|}{|D_{val}|} \times \hat{\sigma}^l_k
\end{equation}
where, $|G_k|$ is the number of inputs in group $k$ and $|D_{val}|$ is the number of inputs in validation set.
For example, if all the inputs belong to $G_k$, the temperature parameter from the temperature scaling will be used, and if there is no inputs in $G_k$, the temperature parameter is equal to 1, which means no calibration will be applied to $G_k$. 
With these modified temperature parameters, both original inputs logits and transformed inputs logits are computed.
These temperature parameters and updated logits are stored for the test time and the later iterations, respectively.

\subsection{Runtime Confidence Calculation using ReCal}
\vspace{-.5em}
After the calibration on validation set finished, inputs in test set can be calibrated as described in Algorithm \ref{alg:apply_cal}.
Because the transformation pool and a transformation at each iteration are already prepared in the calibration step, applying calibration step starts with computing base logits.
After the base logits of original inputs and transformation inputs are ready, the iterative procedures will be repeated for $l$ iterations, as determined in the calibration step.

\subsection{Limitations of ReCal}
\vspace{-.5em}
ReCal has a few limitations.
First, we have to have a label-invariant transformation which lose information, which is necessary for lossy label-invariant grouping step.
However, it is not hard to find such transformations.
For example, in image classifications, besides zoom-out transformation and brightness transformation used in this paper, image blurring / pixalization, lossy compression, and random pixel changes are other possible examples. 
Regarding classification on time-series data such as video and medical data, introducing missing data is a type of lossy label-invariant transformation since the act of losing a frame or occasional data sample doesn't change the state of the environment or patient.

Next, ReCal needs to consider $N$ possible transformed inputs which may be inefficient memory-wise.
However, what we mainly use is the logits not the inputs, since once we compute the base logits at the beginning, we do not use the inputs anymore.
This logits are much smaller compared to the inputs because the logits size is equal to the number of classes and in classification task.
For example, each ImageNet input data has a dimension of $224 \times 224 \times 3$ which takes about $588$ KB, while the logits have a dimension of $1,000$, which is about $3.9$ KB. 

\section{Experiments}
\label{sec:experiments}
\vspace{-1em}

We apply ReCal to multiple models on three datasets to compare its calibration performance and scalability.
In detail, we train or obtain models for each dataset, and calibrate confidence using ReCal and other baselines.
We then compare the calibration performance using two metrics and the time for learning a calibration map to evaluate the scalability.
The details of datasets, model, baselines, metrics and results are described in the following subsections.

\subsection{Experimental Setup}
\vspace{-.5em}
This subsection will explain the datasets, models, baselines, and evaluation metrics for the experiments.
In detail, the first subsection briefly describes datasets such as the number of classes and images, followed by the section for models; which models are used for each dataset and how we obtain the models.
The next subsection is for describing what other calibration algorithms is used as baselines, and the final subsection illustrates the evaluation metrics.

\textbf{Datasets and Models.}
We perform experiments on three datasets: CIFAR10, CIFAR100 \citep{krizhevsky2009learning}, and ImageNet \citep{deng2009imagenet}.  For CIFAR10/100, we use DenseNet40 \citep{huang2017densely}, LeNet5 \citep{lecun1998gradient}, ResNet110 \citep{he2016deep}, ResNet110 SD \citep{huang2016deep}, and WRN-28-10 \citep{zagoruyko2016wide}. For ImageNet, we use DenseNet161 \citep{huang2017densely} and ResNet152 \citep{he2016deep}.  Complete details of the datasets and models are provided in the supplementary material.

\textbf{Competing Approaches for Baseline Comparisons.}
We compare ReCal with various other calibration methods such as temperature scaling, vector scaling, MS-ODIR, Dir-ODIR.
Among those methods, temperature scaling keeps the original accuracy, and other methods change the accuracy.
We implement temperature scaling, and obtain codes for vector scaling, MS-ODIR, and Dir-ODIR from the paper's repository \citep{kull2019beyond}.

\textbf{Evaluation Metrics.}
Our main goal is minimizing the worst-case confidence error, however, as described in Section \ref{sec:problem_statement}, it is impossible to quantify due to the absence of available datasets with confidence estimates for the label.
Instead, we aim to minimize ECE, and our main evaluation metric for the experiments is ECE.
Besides ECE, we also compare approaches using Brier score \citep{brier1950verification}, which considers accuracy as well.  For completeness, definitions of ECE and Brier score are provided in the supplementary material.
Lastly, for assessing the scalability, we compute the learning time of a calibration map.
\begin{table*}[!bt]
\caption{ECE}
\label{tab:exp_res_ECE}
\begin{center}
\resizebox{\textwidth}{!}{\begin{tabular}{c|c|c|c|c|c|c|c|c|c}
Dataset & Model & Uncal. & TS & VS & MS-ODIR & Dir-ODIR & \thead{ReCal\\(z, .1-.9, 20)} & \thead{ReCal\\(z, .5-.9, 10)} & \thead{ReCal\\(b, .1-.9, 20)}\\
\hline
\hline
CIFAR10 & DenseNet40 & 0.052026 & 0.007037 & \underline{0.004438} & 0.005161 & \textbf{0.003943} & 0.010143 & 0.008721 & 0.005892\\
CIFAR10 & LeNet5 & 0.018170 & 0.011963 & \textbf{0.009174} & 0.014147 & 0.010525 & 0.011785 & \underline{0.010507} & 0.010669\\
CIFAR10 & ResNet110 & 0.045646 & 0.008770 & 0.009442 & 0.008829 & \underline{0.008366} & 0.008986 & \textbf{0.008206} & 0.009177\\
CIFAR10 & ResNet110 SD & 0.053770 & 0.011407 & \textbf{0.008552} & 0.010187 & \underline{0.009369} & 0.011973 & 0.012103 & 0.012845\\
CIFAR10 & WRN 28-10 & 0.025076 & 0.009709 & 0.009564 & \underline{0.009175} & 0.009429 & \textbf{0.009092} & 0.012459 & 0.010261\\
\hline
CIFAR100 & DenseNet40 & 0.172838 & 0.015435 & 0.026634 & 0.029628 & 0.018949 & \underline{0.015398} & \textbf{0.011713} & 0.018059\\
CIFAR100 & LeNet5 & \textbf{0.009991} & 0.021064 & 0.015524 & \underline{0.013149} & 0.014172 & 0.019196 & 0.018426 & 0.019367\\
CIFAR100 & ResNet110 & 0.142223 & \underline{0.009101} & 0.029982 & 0.034519 & 0.023109 & 0.012142 & \textbf{0.008487} & 0.010614\\
CIFAR100 & ResNet110 SD & 0.122932 & \underline{0.009310} & 0.035832 & 0.035478 & 0.020747 & 0.009987 & 0.014375 & \textbf{0.007918}\\
CIFAR100 & WRN 28-10 & 0.053396 & 0.043703 & 0.045178 & 0.035509 & \textbf{0.034604} & 0.037270 & \underline{0.035279} & 0.035435\\
\hline
ImageNet & DenseNet161 & 0.056384 & 0.019873 & 0.023286 & 0.036785 & 0.047707 & \textbf{0.013348} & \underline{0.014474} & 0.016981\\
ImageNet & ResNet152 & 0.049142 & 0.020069 & 0.020672 & 0.034736 & 0.039748 & \underline{0.013869} & \textbf{0.013491} & 0.017483\\
\end{tabular}
}
\end{center}
\end{table*}

\begin{table*}[!bt]
\caption{Brier Score}
\label{tab:exp_res_Brier}
\begin{center}
\resizebox{\textwidth}{!}{\begin{tabular}{c|c|c|c|c|c|c|c|c|c}
Dataset & Model & Uncal. & TS & VS & MS-ODIR & Dir-ODIR & \thead{ReCal\\(z, .1-.9, 20)} & \thead{ReCal\\(z, .5-.9, 10)} & \thead{ReCal\\(b, .1-.9, 20)}\\
\hline
\hline
CIFAR10 & DenseNet40 & 0.013585 & 0.012330 & 0.012300 & 0.012256 & 0.012296 & \textbf{0.012225} & \underline{0.012231} & 0.012324\\
CIFAR10 & LeNet5 & 0.037836 & 0.037792 & 0.037748 & 0.037745 & 0.037706 & \textbf{0.037395} & \underline{0.037403} & 0.037784\\
CIFAR10 & ResNet110 & 0.011537 & 0.010439 & 0.010378 & 0.010382 & 0.010350 & \underline{0.010322} & \textbf{0.010317} & 0.010441\\
CIFAR10 & ResNet110 SD & 0.015472 & 0.014395 & 0.014325 & 0.014231 & 0.014302 & \underline{0.014212} & \textbf{0.014140} & 0.014425\\
CIFAR10 & WRN 28-10 & 0.006731 & 0.006357 & 0.006380 & 0.006342 & \underline{0.006336} & \textbf{0.006300} & 0.006344 & 0.006363\\
\hline
CIFAR100 & DenseNet40 & 0.004862 & 0.004329 & 0.004346 & 0.004333 & 0.004318 & \underline{0.004304} & \textbf{0.004302} & 0.004332\\
CIFAR100 & LeNet5 & 0.007581 & 0.007588 & 0.007587 & 0.007580 & 0.007567 & \underline{0.007557} & \textbf{0.007543} & 0.007581\\
CIFAR100 & ResNet110 & 0.004521 & 0.004144 & 0.004180 & 0.004178 & 0.004149 & \underline{0.004130} & \textbf{0.004119} & 0.004149\\
CIFAR100 & ResNet110 SD & 0.004344 & 0.004064 & 0.004046 & 0.004045 & 0.004047 & \underline{0.004035} & \textbf{0.004028} & 0.004067\\
CIFAR100 & WRN 28-10 & 0.002929 & 0.002915 & 0.002948 & \underline{0.002901} & \textbf{0.002898} & 0.002913 & 0.002913 & 0.002926\\
\hline
ImageNet & DenseNet161 & 0.000323 & 0.000319 & \underline{0.000316} & \textbf{0.000313} & 0.000324 & 0.000318 & 0.000319 & 0.000319\\
ImageNet & ResNet152 & 0.000305 & 0.000302 & \underline{0.000301} & \textbf{0.000299} & 0.000307 & 0.000302 & 0.000302 & 0.000302\\
\end{tabular}
}
\end{center}
\end{table*}

\subsection{Results}
\vspace{-.5em}
We analyze the results in terms of calibration performance and time for learning a calibration map.
First, we compare the calibration performance in terms of ECE and brier score.
ECE is for evaluating how well each algorithms calibrate confidence.
Brier score is for the similar evaluation, but, this metric considers the prediction accuracy together.
Second, we present the time for learning a calibration map so that we assess the scalability.
\subsubsection{Calibration Performance.}
\vspace{-.5em}
We display the calibration performance of various methods in Table \ref{tab:exp_res_ECE} and \ref{tab:exp_res_Brier}.
Table \ref{tab:exp_res_ECE} and \ref{tab:exp_res_Brier} display ECE and Brier score, and test error rates is shown in supplementary material.
The values with bold and with underline represent the best and the second best result, respectively.

For ReCal, we show the three different transformation pool: (z, .1-.9, 20), (z, .5-.9, 10), (b, .1-.9, 20).
The first parameter means the transformation type; z and b mean zoom-out transformation and brightness transformation, respectively.
Next parameter represents the range of transformation parameters, the range is either from 0.1 to 0.9 or from 0.5 to 0.9.
The last parameter corresponds to the number of transformation.
We use 20 transformations when we have the range of from 0.1 to 0.9, and 10 transformations for the range of from 0.5 to 0.9.

\emph{\textbf{ECE results.}}
Table \ref{tab:exp_res_ECE} shows ECE values of all datasets and models.
For CIFAR10, vector scaling, Dir-ODIR, and ReCal shows the best performance on 2/1/2 models, respectively.
For CIFAR100, except LeNet5 and WRN-28-1, ReCal shows the best performance.
Among our methods, for most cases, (z, .5-.9, 10) shows the best performance.
For ImageNet, ReCal has the best ECE for both of models; specifically, (z, .1-.9, 20) and (z, .5, .9, 20) are the best for each model.

\emph{\textbf{Brier score results.}}
Brier scores are displayed in Table \ref{tab:exp_res_Brier}.
For CIFAR10/100, ReCal almost always shows the best performance.
The only exception is when Dir-ODIR is applied to WRN 28-10 on CIFAR100.
For ImageNet, MS-ODIR shows the best performance and vector scaling shows the second-best value.
ReCal is slightly higher than those values.
The reason that ReCal shows worse Brier score compared to vector scaling and MS-ODIR is that those two calibration methods increase the accuracy.
Brier score considers both of accuracy and calibration, and due to the increase of accuracy on ImageNet results in the better Brier score.

\emph{\textbf{Comparison between ReCal settings.}}
We show the three different settings of ReCal; two different transformations, and two different transformation parameter ranges for zoom-out transformation.
We compare these three settings in terms of two aspects.
First, between brightness and zoom-out transformation, zoom-out calibrates better, especially, on ImageNet.
We conjecture that the reason is related to the fact that zoom-out is more effective in Lossy Label-Invariant Grouping as described in Section \ref{sec:group_input}.
Next, for zoom-out transformations, the appropriate parameter range is related to the original image resolution.
Specifically, a scale factor range of between 0.5 and 0.9 generally shows better performance on CIFAR10/100, and a scale factor range of 0.1 and 0.9 is better on ImageNet based on ECE and Brier score. Therefore, we suggest to use the small zoom-out scale factors for only large images. %

\subsubsection{Learning Time}  
\vspace{-.5em}
We also compute the learning time of each calibration algorithms on various datasets and models, and the result is shown in supplementary materials.

Temperature scaling is generally the fastest algorithm, and the next order is vector scaling, our method, Dir-ODIR, and MS-ODIR.
Because MS-ODIR and Dir-ODIR train multiple calibration models to search the optimal hyper-parameters, its calibration time is high compared to other methods.
Those two algorithms train less number of calibration models for CIFAR100 compared to CIFAR10.
Similarly, we reduce the number of calibration models further for ImageNet, since it is larger than the two datasets.

For ImageNet, our method takes about 51,000 seconds, or 14.1 hours for DenseNet161 and 71,000 seconds, or 19.8 hours for ResNet152.
Even though this is slower than other methods like temperature scaling, and vector scaling, we think that our method can be applied to ImageNet in terms of learning time.
The most slowest time is 380,000 seconds, or 4.4 days for DenseNet161, and 220,000 seconds, or 2.5 days for ResNet152.

\vspace{-1em}
\section{Conclusion}
\label{sec:conclusion}
\vspace{-1em}

In this paper, we propose an accuracy preserving post-hoc calibration method based on a label-invariant image transformation.
ReCal exploits the properties of label-invariant  transformations to group inputs,
and applies different temperature scaling to each group.
Because ReCal is based on temperature scaling, it preserves the original classifier accuracy.
In addition, it has more expressiveness compared to original temperature scaling because it uses multiple temperature scaling coefficients.
Experiments on CIFAR10/100 and ImageNet
datasets show that ReCal can be applied to the large-scale ImageNet, and outperforms other methods on those dataset including ImageNet.

For the future work, incorporating multiple types of transformation type may improve the calibration performance. 
In this paper, we use one type of transformation at a time, but, different transformation utilizes different information in example space, and combination of multiple transformation type may results in improvements.
Additionally, ReCal can be extended to other types of dataset as long as appropriate transformation exists.
For example, we can apply ReCal to time-series data classification. We can consider a transformation of eliminating some data at random time point. This transformation is a label-invariant transformation which decrease confidence, and ReCal can be applied to calibrate confidence.

\section*{Acknowledgement}
\label{acknowledgement}
This work was supported by the Air Force Research Laboratory and the Defense Advanced Research Projects Agency under Contract No. FA8750-18-C-0090, and by the Army Research Office under Grant Number W911NF-20-1-0080. Any opinions, findings and conclusions or recommendations expressed in this material are those of the authors and do not necessarily reflect the views of the Air Force Research Laboratory (AFRL), the Army Research Office (ARO), the Defense Advanced Research Projects Agency (DARPA), or the Department of Defense, or the United States Government.

\bibliography{main-arxiv.bbl}
\setcounter{section}{0}
\renewcommand\thesection{\Alph{section}}

\onecolumn
\title{Improving Classifier Confidence using Lossy Label-Invariant Transformations: \\
Supplementary Materials}

\section{Brier score}
Brier score \citep{brier1950verification} is defined as Equation \ref{eq:brier_def}.
\begin{equation}
\label{eq:brier_def}
    Brier = \frac{1}{N}\sum_{i=1}^N\sum_{k=1}^K(p_{ik} -y_{ik})^2
\end{equation}
where, $N$ is the dataset size, $K$ is the number of classes, $p_{ik}$ is the confidence for the label k of the $i^{th}$ data, and $y_{ik}$ is 1 if the true label for $i^{th}$ data is k otherwise 0. 
Here, we normalized Brier score by dividing it by the number of classes as used in \cite{kull2019beyond}.
The formal definition of the normalized Brier score is shown in Equation \ref{eq:norm_brier_def}.

\begin{equation}
\label{eq:norm_brier_def}
\textit{Normalized Brier} = \frac{1}{NK}\sum_{i=1}^N\sum_{k=1}^K(p_{ik} -y_{ik})^2
\end{equation}

\section{Flowchart for Runtime Confidence Calculation using ReCal}
\begin{figure*}[!hbt]
    \centerline{
    \includegraphics[width=1\textwidth]{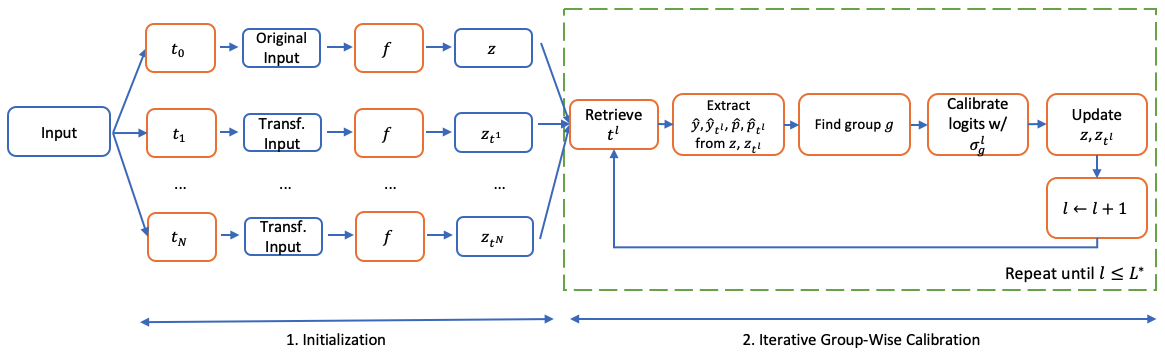}
    }
    \caption{Runtime Confidence Calculation using ReCal}
    \label{fig:flow_chart_apply}
\end{figure*}

After ReCal learns a calibration map, it can process runtime calibration as illustrated in Figure \ref{fig:flow_chart_apply}.
It consists of two steps: initialization step, iterative group-wise calibration.
In initialization step, it computes the base logits for original inputs and transformed inputs by the sampled transformation during the calibration process.
After the initialization step, it repeats 1) Find a group number for the given input, 2) Calibrate logits using given temperature parameter.
The detail explanation about this process is in Section \ref{sec:algorithm}.

\section{Additional Results}
In this section, we display more results for Section \ref{sec:group_input} and Section \ref{sec:experiments}.
For Section \ref{sec:group_input}, we present the detail ECE and number of images of each group for each transformation type and parameter, and for Section \ref{sec:experiments}, we show the test error rate and learning time of a calibration map.

\subsection{Detail Result for Lossy Label-Invariant Grouping}
Table \ref{tab:grp_img_resnet152_zoom} and \ref{tab:grp_img_resnet152_brightness} show ECE and number of images of each group for different parameters using zoom-out transformation and brightness transformation.
The second row represents ECE and number of images of test data, and following rows show the ECE and the number of images of each group for each transformation parameter.
The bold and italic numbers mean the best and worst result among four groups, respectively.

As shown in Table \ref{tab:grp_img_resnet152_zoom}, with zoom-out transformation, group 4 has the best ECE, \ie this group requires less calibration and group 1 has the worst ECE \ie this group requires more calibration, and group 2 and 3 have the medium range of ECE, for the most of transformation parameters except 0.1x and 0.2x.
The similar pattern with more variability is observed with brightness transformation as displayed in Table \ref{tab:grp_img_resnet152_brightness}.
With brightness transformation, group 4 has the best ECE for the transformation less than 0.8x, and group 1 has the worst ECE for the transformation less than 0.5x.
Unlike zoom-out transformation, group 2 also have the worst ECE for about the half of transformation parameters.
However, each group still show different ECE compared to other groups, which supports our idea of group-wise calibration.

Table \ref{tab:grp_img_resnet152_zoom} and \ref{tab:grp_img_resnet152_brightness} show the number of images in each group.
For different parameters, the image distribution over groups are different, and zoom-out transformation shows more variability than brightness transformation.
Based on this observation, we design an ReCal which can incorporate multiple parameters as described in Section \ref{sec:algorithm}.
Lastly, for zoom-out transformation with the scale factor of 0.1x, group 3 has only three images.
With this small amount of images, calibration can overfit the data, and we address this issue as described in Section \ref{sec:algorithm}.

\begin{table}[!bt]
\caption{Grouping Image Using Zoom-Out Transformation}
\label{tab:grp_img_resnet152_zoom}
\begin{center}
\begin{tabular}{c|c|c|c||c|c}
& &  \multicolumn{2}{c||}{ECE} & \multicolumn{2}{c}{Count}\\
\hline
& Test Data & \multicolumn{2}{c||}{0.020069} & \multicolumn{2}{c}{25000}\\
\hline
& & Incr. & Not Incr.& Incr. & Not Incr.\\
\hline
\multirow{2}{*}{0.9x} & Change & \emph{0.047142} & 0.040512 & 1578 & 2328\\
& No Change & 0.025389 & \textbf{0.020825} & 8250 & 12844\\
\hline
\multirow{2}{*}{0.8x} & Change & \emph{0.048417} & 0.033766 & 1830 & 3084\\
& No Change & 0.025770 & \textbf{0.020072} & 7038 & 13048\\
\hline
\multirow{2}{*}{0.7x} & Change & \emph{0.036925} & 0.034266 & 1938 & 4149\\
& No Change & 0.029248 & \textbf{0.019347} & 6109 & 12804\\
\hline
\multirow{2}{*}{0.67x} & Change & \emph{0.040322} & 0.032636 & 1990 & 4690\\
& No Change & 0.028604 & \textbf{0.018290} & 5507 & 12813\\
\hline
\multirow{2}{*}{0.5x} & Change & \emph{0.044078} & 0.026399 & 2272 & 8444\\
& No Change & 0.028710 & \textbf{0.016006} & 3744 & 10540\\
\hline
\multirow{2}{*}{0.4x} & Change & \emph{0.044421} & 0.022210 & 2041 & 12630\\
& No Change & 0.034228 & \textbf{0.016316} & 2096 & 8233\\
\hline
\multirow{2}{*}{0.33x} & Change & \emph{0.054198} & 0.019901 & 1837 & 16635\\
& No Change & 0.038278 & \textbf{0.017077} & 1067 & 5461\\
\hline
\multirow{2}{*}{0.2x} & Change & 0.073773 & \textbf{0.019371} & 791 & 23446\\
& No Change & \emph{0.205023} & 0.030856 & 58 & 705\\
\hline
\multirow{2}{*}{0.1x} & Change & 0.069130 & \textbf{0.019659} & 344 & 24596\\
& No Change & \emph{0.161533} & 0.098585 & 3 & 57\\
\end{tabular}

\end{center}
\end{table}

\begin{table}[!bt]
\caption{Grouping Image Using Brightness}
\label{tab:grp_img_resnet152_brightness}
\begin{center}
\begin{tabular}{c|c|c|c||c|c}
& &  \multicolumn{2}{c||}{ECE} & \multicolumn{2}{c}{Count}\\
\hline
& Test Data & \multicolumn{2}{c||}{0.020069} & \multicolumn{2}{c}{25000}\\
\hline
& & Incr. & Not Incr.& Incr. & Not Incr.\\
\hline
\multirow{2}{*}{0.9x} & Change & 0.066956 & \emph{0.077188} & 315 & 356\\
& No Change & \textbf{0.021815} & 0.022545 & 11404 & 12925\\
\hline
\multirow{2}{*}{0.8x} & Change & 0.059497 & \emph{0.073143} & 556 & 625\\
& No Change & \textbf{0.023280} & 0.023908 & 11020 & 12799\\
\hline
\multirow{2}{*}{0.7x} & Change & 0.041225 & \emph{0.052961} & 750 & 951\\
& No Change & 0.025298 & \textbf{0.022240} & 10623 & 12676\\
\hline
\multirow{2}{*}{0.67x} & Change & 0.039053 & \emph{0.061532} & 817 & 1060\\
& No Change & 0.024897 & \textbf{0.022962} & 10437 & 12686\\
\hline
\multirow{2}{*}{0.5x} & Change & 0.046058 & \emph{0.053280} & 1172 & 1583\\
& No Change & 0.026930 & \textbf{0.022807} & 9355 & 12890\\
\hline
\multirow{2}{*}{0.4x} & Change & \emph{0.045266} & 0.040179 & 1332 & 2109\\
& No Change & 0.027172 & \textbf{0.022091} & 8462 & 13097\\
\hline
\multirow{2}{*}{0.33x} & Change & \emph{0.048402} & 0.042858 & 1446 & 2572\\
& No Change & 0.025383 & \textbf{0.023382} & 7676 & 13306\\
\hline
\multirow{2}{*}{0.2x} & Change & \emph{0.052800} & 0.035715 & 1578 & 4337\\
& No Change & 0.027635 & \textbf{0.017540} & 5394 & 13691\\
\hline
\multirow{2}{*}{0.1x} & Change & \emph{0.040800} & 0.029543 & 1198 & 9341\\
& No Change & 0.034183 & \textbf{0.016133} & 2360 & 12101\\
\end{tabular}
\end{center}
\end{table}

\subsection{Additional Comparison Results}

Besides ECE and Brier Score, we also compare test error rate and learning time of a calibration map.
Table \ref{tab:exp_res_Error} and \ref{tab:exp_res_learn_time} show the test error rate, and the learning time, respectively.
For tables, bold numbers mean the best results and underlined numbers represent the second-best results.

\paragraph{Error rate.}
We calculate test error rate to compare the accuracy preserving properties.
As shown in Table \ref{tab:exp_res_Error}, temperature scaling (TS), ReCal do not change the original accuracy, \ie their error rate is equal to an uncalibrated classifier's one.
However, vector scaling (VS), MS-ODIR, and Dir-ODIR change the original accuracy.
Without a consistent pattern, all of those calibration algorithms increase or decrease the error rate depending on the dataset and model.
In detail, vector scaling decreases the original classifier's accuracy except the ImageNet experiments.
MS-ODIR hurts the original classifier's accuracy except for DenseNet40 and ResNet110 SD on CIFAR10 and DenseNet161, ResNet152 on ImageNet.
Dir-ODIR worsen the original classifier's accuracy except for  LeNet5 on CIFAR10, LeNet5/Resnet110/ResNet110 SD on CIFAR100.

\begin{table*}[!bt]
\caption{Test Error Rate (\%)}
\label{tab:exp_res_Error}
\begin{center}
\resizebox{\textwidth}{!}{\begin{tabular}{c|c|c|c|c|c|c|c|c|c}
Dataset & Model & Uncal. & TS & VS & MS-ODIR & Dir-ODIR & \thead{ReCal\\(z, .1-.9, 20)} & \thead{ReCal\\(z, .5-.9, 10)} & \thead{ReCal\\(b, .1-.9, 20)}\\
\hline
\hline
CIFAR10 & DenseNet40 & \underline{8.25} & \underline{8.25} & 8.31 & \textbf{8.22} & 8.31 & \underline{8.25} & \underline{8.25} & \underline{8.25}\\
CIFAR10 & LeNet5 & \underline{27.23} & \underline{27.23} & 27.33 & 27.24 & \textbf{27.20} & \underline{27.23} & \underline{27.23} & \underline{27.23}\\
CIFAR10 & ResNet110 & \textbf{6.90} & \textbf{6.90} & 7.06 & 7.06 & \underline{7.03} & \textbf{6.90} & \textbf{6.90} & \textbf{6.90}\\
CIFAR10 & ResNet110 SD & \underline{9.62} & \underline{9.62} & 9.76 & \textbf{9.59} & 9.64 & \underline{9.62} & \underline{9.62} & \underline{9.62}\\
CIFAR10 & WRN 28-10 & \textbf{4.06} & \textbf{4.06} & \underline{4.10} & 4.13 & \underline{4.10} & \textbf{4.06} & \textbf{4.06} & \textbf{4.06}\\
\hline
CIFAR100 & DenseNet40 & \textbf{31.84} & \textbf{31.84} & 32.27 & 32.00 & \underline{31.89} & \textbf{31.84} & \textbf{31.84} & \textbf{31.84}\\
CIFAR100 & LeNet5 & \underline{62.34} & \underline{62.34} & 62.66 & 62.58 & \textbf{62.22} & \underline{62.34} & \underline{62.34} & \underline{62.34}\\
CIFAR100 & ResNet110 & \underline{30.48} & \underline{30.48} & 30.94 & 30.80 & \textbf{30.46} & \underline{30.48} & \underline{30.48} & \underline{30.48}\\
CIFAR100 & ResNet110 SD & \underline{29.90} & \underline{29.90} & 29.98 & 29.91 & \textbf{29.89} & \underline{29.90} & \underline{29.90} & \underline{29.90}\\
CIFAR100 & WRN 28-10 & \textbf{20.10} & \textbf{20.10} & \underline{20.29} & 20.47 & 20.51 & \textbf{20.10} & \textbf{20.10} & \textbf{20.10}\\
\hline
ImageNet & DenseNet161 & 22.55 & 22.55 & \underline{22.49} & \textbf{22.10} & 23.07 & 22.55 & 22.55 & 22.55\\
ImageNet & ResNet152 & 21.31 & 21.31 & \underline{21.22} & \textbf{20.96} & 21.63 & 21.31 & 21.31 & 21.31\\
\end{tabular}
}
\end{center}
\end{table*}

\paragraph{Learning Time.}
We display the learning time of a calibration map for each algorithm in Table \ref{tab:exp_res_learn_time}.
Temperature scaling (TS) is always the fastest calibration algorithm followed by vector scaling (VS).
The next fastest one is ReCal, and we think ReCal can be applied to ImageNet in terms of the learning time.
Specifically, it takes 50,730 seconds or 14.1 hours for DenseNet161 on ImageNet and 71,254 seconds or 19.8 hours for ResNet152 on ImageNet.
On the other hand, MS-ODIR and Dir-ODIR are slower than other calibration algorithms because it basically calibrate many models to find appropriate its hyper-parameters.

\begin{table*}[!hbt]
\caption{Learning Time (sec)}
\label{tab:exp_res_learn_time}
\begin{center}
\begin{tabular}{c|c|c|c|c|c|c}
Dataset & Model & TS & VS & MS-ODIR & Dir-ODIR & \thead{ReCal\\(z, .1-.9, 20)} \\
\hline
\hline
CIFAR10 & DenseNet40 & \textbf{2.94} & \underline{31.10} & 77353.63 & 43001.99 & 84.04\\
CIFAR10 & LeNet5 & \textbf{1.86} & \underline{12.06} & 42830.58 & 37001.63 & 110.79\\
CIFAR10 & ResNet110 & \textbf{2.21} & \underline{26.65} & 70702.87 & 45836.87 & 38.85\\
CIFAR10 & ResNet110 SD & \textbf{4.35} & \underline{26.52} & 85859.16 & 54783.42 & 58.74\\
CIFAR10 & WRN 28-10 & \textbf{7.68} & \underline{28.22} & 67955.20 & 36386.26 & 49.62\\
\hline
CIFAR100 & DenseNet40 & \textbf{14.03} & \underline{26.31} & 320284.77 & 134317.54 & 136.23\\
CIFAR100 & LeNet5 & \textbf{9.63} & \underline{26.10} & 109645.75 & 83324.48 & 97.77 \\
CIFAR100 & ResNet110 & \textbf{8.63} & \underline{26.61} & 300360.19 & 134317.54 & 97.29\\
CIFAR100 & ResNet110 SD & \textbf{13.24} & \underline{26.73} & 276767.31 & 126100.97 & 604.12\\
CIFAR100 & WRN 28-10 & \textbf{14.23} & \underline{25.60} & 161327.35 & 85532.50 & 125.84\\
\hline
ImageNet & DenseNet161 & \textbf{865.40} & \underline{285.73} & 379487.45 & 276553.98 & 50730.17\\
ImageNet & ResNet152 & \textbf{754.51} & \underline{342.50} & 215746.16 & 229493.41 & 71254.34\\
\end{tabular}

\end{center}
\end{table*}




\end{document}